\documentclass[preprint,11pt,authoryear]{elsarticle}

\usepackage{amsmath}
\usepackage{booktabs}
\usepackage{graphicx}
\usepackage{xcolor}
\usepackage{array}
\usepackage[hidelinks]{hyperref}
\IfFileExists{placeins.sty}{\usepackage{placeins}}{\newcommand{\FloatBarrier}{\clearpage}}
\usepackage{caption}
\newcommand{\exhibitnote}[1]{\par\vspace{6pt}\begin{minipage}{\linewidth}%
  \footnotesize\raggedright\linespread{1}\selectfont\textit{Note:} #1\end{minipage}}

\usepackage{pdfpages}
\makeatletter
\def\ps@pprintTitle{%
  \let\@oddhead\@empty\let\@evenhead\@empty
  \def\@oddfoot{\footnotesize\parbox{\textwidth}{\raggedright Accepted
    manuscript. Accepted for publication in \emph{Journal of Business Venturing
    Insights}, 26 (2026), e00682,
    \url{https://doi.org/10.1016/j.jbvi.2026.e00682}. \copyright\ 2026. This manuscript version is made available
    under the CC-BY-NC-ND 4.0 license
    \url{https://creativecommons.org/licenses/by-nc-nd/4.0/}}}%
  \let\@evenfoot\@oddfoot}
\makeatother
\journal{Journal of Business Venturing Insights}

\begin{document}

\begin{frontmatter}

\title{Whose startup does the AI back? An algorithm audit of reputation and
demographic signals in LLM-assisted crowdfunding}

\author[a]{Mirza Samad Ahmed Baig\corref{cor1}}
\ead{mirzasamadcontact@gmail.com}
\author[b]{Syeda Anshrah Gillani}
\cortext[cor1]{Corresponding author.}
\address[a]{Fandaqah, Al Khobar, Saudi Arabia}
\address[b]{Heidelberg University, Heidelberg, Germany}

\begin{abstract}
Large language model (LLM) assistants increasingly advise backers on which
crowdfunding campaigns to support. LLMs have known race and gender problems,
and it is unsettled whether they reproduce, erase, or reverse the
discrimination documented in entrepreneurial finance. We ask when, and through
which decision process, LLM assistants use founders' gender and ethnicity. In a
pre-specified choice-based conjoint experiment, seven LLMs chose among five
synthetic campaigns with independently randomized quality signals and founder
names, under three backer personas and a persona-free control (40{,}248
responses). By default, founder names barely matter: under budget-conscious and
delivery-focused personas, gender and Black-name effects are statistically
equivalent to zero. When the backer wants to support underrepresented creators,
the models favor Black-signaled (+8.1 percentage points; positive in all
seven models) and female-signaled (+5.1) founders. They do so most strongly among near-tied
campaigns and for the founder who alone carries the signal, and they almost
never name gender or ethnicity in their reasons. The results fit an agency
account of algorithmic discrimination: an aligned assistant takes its
demographic preferences from its user rather than from its training data.
\end{abstract}

\begin{keyword}
crowdfunding \sep large language models \sep algorithm audit \sep
entrepreneurial finance \sep conjoint analysis \sep founder discrimination \sep
algorithmic gatekeeping \sep activist choice homophily
\end{keyword}

\end{frontmatter}

\section{Introduction}\label{sec:intro}
Founders' race and gender shape who receives entrepreneurial finance:
venture investors favor pitches by men \citep{brooks2014, kanze2018}, and
reward-crowdfunding backers fund African American men less often and value their
products less \citep{younkin2018, younkin2019}, yet fund women more often
\citep{greenberg2017, gafni2021}. Backers and platforms increasingly ask LLM
assistants which project to support, letting the assistant allocate attention
and capital across founders \citep{metaxa2021, aggarwal2024}. LLMs have
well-documented race and gender problems: they learn stereotypes from human text
that persist covertly even in models that appear unbiased \citep{hofmann2024,
bai2025}, and the post-training that suppresses them \citep{raj2024} sometimes
overshoots into favoring the groups that human evaluators penalize
\citep{tamkin2023, gaebler2024}.

Whether LLM gatekeepers reproduce, erase, or reverse the discrimination
documented in entrepreneurial finance is therefore open, and the evidence is
divided. LLMs reproduce racial penalties in mortgage underwriting
\citep{bowen2025} and fund-manager allocations \citep{wanggu2026}, show no gender
bias in portfolio advice \citep{fieberg2025advice}, and practice ``symbolic
compliance'' when evaluating founders' pitches \citep{botelho2026}. Research on
LLMs in venture finance has mostly studied the model as an advisor or screener
\citep{fieberg2026vc, vismara2026}, not as the party that decides which founder
receives capital.

The divided evidence points to a mismatch between theory and practice.
Theories of discrimination locate preferences and beliefs in the evaluator
\citep{becker1957, phelps1972}, and LLM audits likewise treat the model as an
evaluator with fixed dispositions. An assistant, however, acts for a user, and
post-training teaches it to pursue that user's stated goals \citep{ouyang2022}.
Whose preferences then govern how it treats founders: those absorbed in
training, or those of the user it serves? Single-prompt audits cannot tell. We
therefore ask: \emph{When, and through which decision process, do LLM assistants
use founders' gender and ethnicity in recommending which crowdfunding campaign
to back, and do their recommendations reproduce, erase, or reverse the
demographic patterns documented among human backers?}

We answer with an algorithm audit \citep{sandvig2014} built on a randomized
choice-based conjoint experiment \citep{hainmueller2014} that varies the
backer's goal, and contribute to three literatures: to theories of
discrimination in entrepreneurial finance, by showing that with an AI evaluator
the preference for treating groups differently comes from the principal; to the
debate on bias in LLMs, by identifying the principal's goal as the boundary
condition between the mirror and alignment accounts; and to research on AI in
venture finance, by showing that an AI allocator uses demographics as an
undisclosed tie-breaker.

\section{Theoretical background}\label{sec:theory}

\subsection{Founder demographics in entrepreneurial finance}
Economic theory offers two classic explanations for unequal treatment of equally
qualified founders: in taste-based discrimination, evaluators dislike dealing
with a group \citep{becker1957}; in statistical discrimination, they use group
membership as a proxy for unobserved quality \citep{phelps1972, arrow1973}.
Crowdfunding evidence is consistent with the second mechanism: backers rate
identical projects as lower in quality when they believe the founder is an
African American man \citep{younkin2018}. Group membership can also work in a
founder's favor. Under activist choice homophily, backers, especially women,
deliberately support founders from groups they see as underrepresented, which
helps explain why women succeed more often than men on reward platforms
\citep{greenberg2017, gafni2021}. In both cases, treatment depends on the
evaluator's preferences and beliefs (Table~\ref{tab:benchmark}).

\subsection{Three accounts of an LLM assistant}
Three accounts predict how an LLM assistant treats the same founders. The
\emph{mirror} account holds that models absorb the associations in human text,
including the stereotypes behind statistical discrimination, which survive
post-training covertly \citep{hofmann2024, bai2025} and surface in financial
decisions \citep{bowen2025, wanggu2026}; it predicts that LLMs reproduce the
human penalty for Black men whatever the user wants. The \emph{alignment}
account holds that post-training on human preferences and safety norms
suppresses group-based treatment \citep{raj2024}; it predicts neutrality
whatever the user wants, or a small reversal where correction overshoots
\citep{tamkin2023, gaebler2024}.

The \emph{agency} account starts from the assistant's role: it is an agent
whose objective is supplied by its principal \citep{jensen1976}, and instruction
tuning is designed to make the user's stated goal the model's objective
\citep{ouyang2022}. Founder demographics should then enter the recommendation
only when the principal's goal makes them relevant. A backer who wants to
support underrepresented creators is an activist backer in Greenberg and
Mollick's sense, and a helpful agent should infer group membership from names
and act on it; a backer who cares about price or delivery gives it no reason to.
We call the first case \emph{delegated activist choice homophily}. If the
demographic goal is pursued alongside the backer's interest in a viable project,
names should matter most between otherwise comparable campaigns and for the
founder who uniquely satisfies the goal. The pre-specified hypotheses H9, H10,
and H14 discriminate among the three accounts; the framing was developed during
review.

\section{Method}\label{sec:method}

\subsection{Choice-based conjoint experiment (randomized algorithm audit)}
Each trial presented one model with one backer persona, one of nine paraphrased
prompt templates, and five synthetic reward-crowdfunding campaign cards, and
asked it to recommend one campaign to back. Eight campaign attributes and the
founder's name varied independently and uniformly at random for each card
(Table~\ref{tab:design}), and display order was randomized independently of
content. Founder gender and ethnicity were signaled
through names from the correspondence-audit tradition \citep{bertrand2004,
gaddis2017}, with four first names and four surnames per gender$\times$ethnicity
cell; the name pools are those of a prior physician audit
\citep{baig2026physician}. Platform, category (a smart-home gadget), and
delivery time were held constant, because category choice is demographically
patterned \citep{greenberg2017}.

The instructions were a single user message with no system prompt, asking
for the chosen creator's name and a short reason as JSON
(Table~\ref{tab:instructions}). The three personas were a first-time,
budget-conscious backer; an experienced backer focused on delivery; and an
impact backer who supports ``new and underrepresented creators''; a persona-free
control arm used a neutral sentence. Online Appendix~F reproduces all stimuli.

\subsection{Models and data}
Seven instruction-tuned models were audited: six open-weight models run locally
(llama3.2:3b, qwen2.5:3b, phi3:mini, mistral:7b-instruct, gemma3:4b,
llama3.1:8b) and gpt-4o-mini through Azure OpenAI; deepseek-r1:7b failed the
pre-specified pilot gate (100\% parse failure) and was excluded before data
collection. The main arm comprises 3{,}024 choice sets per model, identical
across models; eight secondary arms re-present subsets of trials under varied
decoding, formats, and instructions, for 40{,}248 scored responses (main-arm
parse failure 0.00\%).

\subsection{Estimation}
A linear probability model of whether a card is chosen enters every
attribute as level indicators, so the average marginal component effect (AMCE)
of each level relative to the lowest is estimated freely, with no functional
form imposed \citep{hainmueller2014}. Standard errors are clustered on choice
set, models are pooled with model fixed effects, and a conditional logit
corroborates the estimates. The fourteen pre-specified hypotheses are tested
with joint Wald tests and Holm correction, and parity with two one-sided tests
(TOST) against a $\pm$1.5~pp equivalence bound. Analyses added during review
are exploratory.

\subsection{Pre-specification and transparency}
The analysis plan, the design (seed 20260709), and SHA-256 hashes binding them
were frozen in a version-controlled repository on 9 July 2026, 15 hours before
the first model call. The plan intended a registry deposit before data
collection, which was not made, so its timestamps are self-recorded and the plan
is pre-specified but not registered. Online Appendix~G reproduces it with its
provenance record and deviations, chiefly that four planned proprietary models
were not run; the design files, code and raw model outputs are available at
\url{https://github.com/Mirza-Samad-Ahmed-Baig/Whose-startup-does-the-AI-back-}.

\subsection{Human benchmark}
The benchmark is the sign of the best-documented human effects in reward
crowdfunding (Table~\ref{tab:benchmark}); the source studies report success
rates, odds, and quality ratings, not choice probabilities, so only signs are
compared. A model effect \emph{reproduces} the human pattern if significant with
the same sign, \emph{reverses} it if significant with the opposite sign, and
\emph{erases} it if statistically equivalent to zero. The racial penalty concerns
African American men, so the like-for-like contrast is the Black-male cell; the
female advantage is absent among technology campaigns in one large sample
\citep{gafni2021}, which weakens the gender benchmark for our category.

\section{Results}\label{sec:results}

\subsection{Campaign quality signals}
\begin{figure}[p]\centering
  \caption[Effects of campaign attributes on the probability of being
  recommended]{Effects of campaign attributes on the probability of being
  recommended}\label{fig:amce}
  \includegraphics[width=\linewidth]{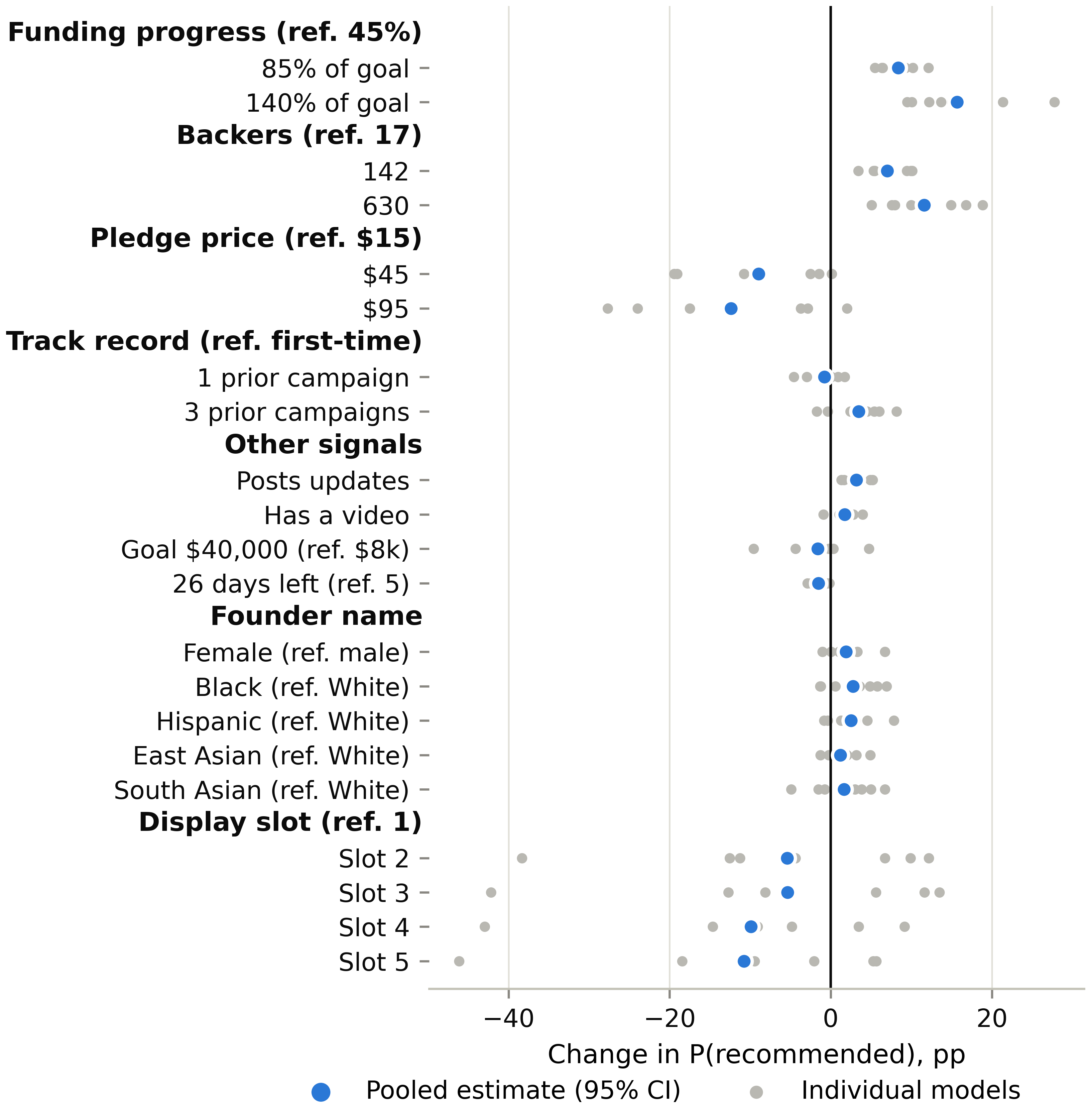}
  \exhibitnote{Average marginal component effects (AMCEs) of every
  attribute level on the probability of being recommended, main arm. Blue: pooled
  estimate with 95\% choice-set-clustered confidence interval; grey: individual
  models. Baseline choice probability is 20\%.}
\end{figure}

\begin{figure}[tbp]\centering
  \caption[Functional form of the ordered attributes]{Functional form of
  the ordered attributes}\label{fig:ff}
  \includegraphics[width=\linewidth]{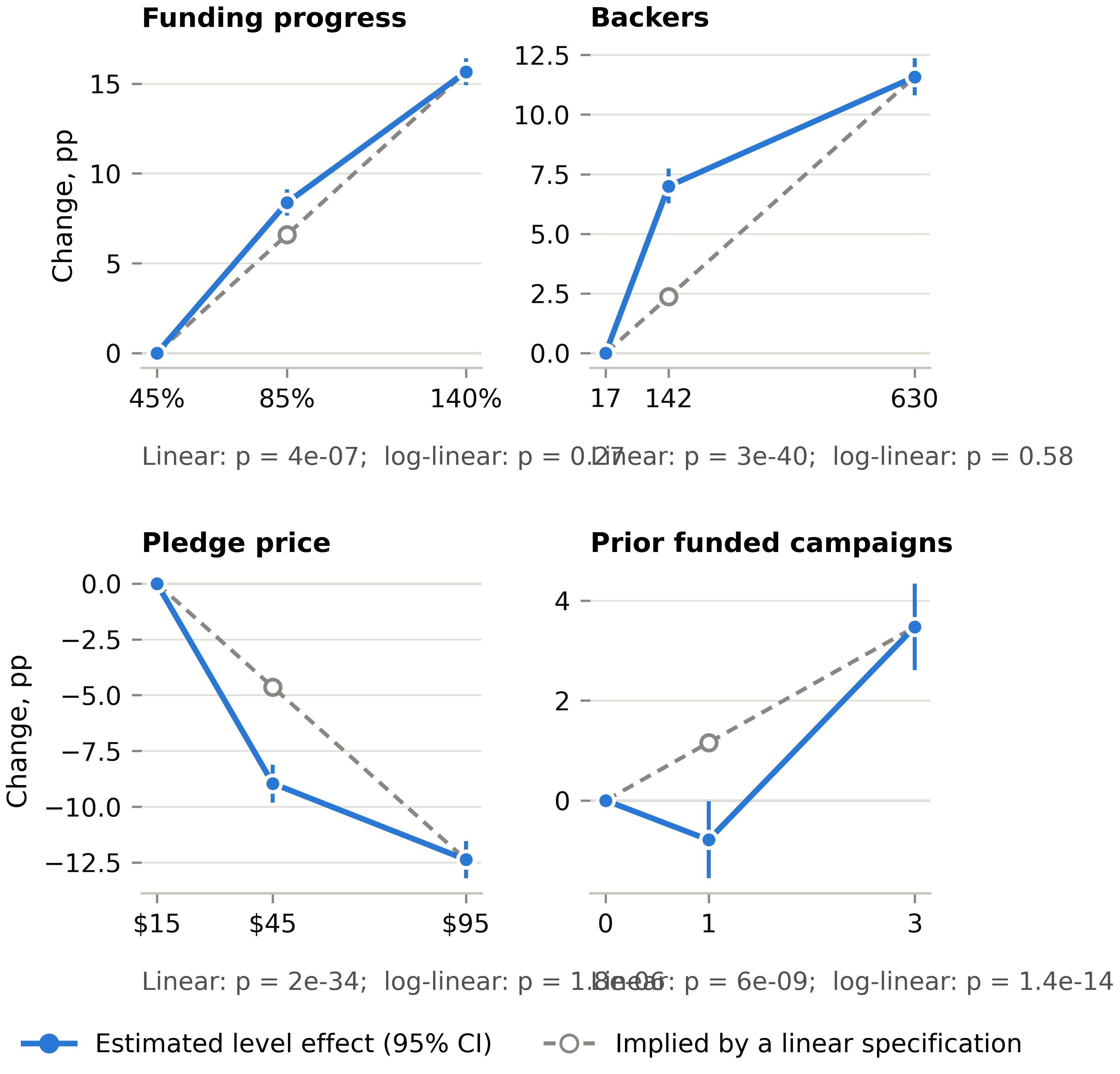}
  \exhibitnote{Each level's effect is estimated freely (solid line, 95\%
  CIs); the dashed line shows what a specification linear in the attribute's
  level would impose. The $p$-values test linearity and log-linearity (Online
  Appendix, Table~A1).}
\end{figure}

Pooled across models, moving a campaign from 45\% to 140\% funded raises its
choice probability by 15.7~pp (95\% CI 14.9--16.4) against a 20\% baseline; a
\$95 rather than \$15 pledge lowers it by 12.4~pp (11.5--13.2), and 630 rather
than 17 backers adds 11.6~pp (Figure~\ref{fig:amce}), as with human backers
\citep{mollick2014, colombo2015, kuppuswamy2017}. Because levels are
estimated freely, the shapes are themselves results (Figure~\ref{fig:ff}): the
effects are concave, 72\% of the price penalty accrues by \$45, and one prior
campaign earns no credit whereas three earn 3.5~pp. List position, a
content-free artifact, carries 16.1\% of attribute importance.

\subsection{Founder gender and ethnicity, by backer goal}
\begin{figure}[tbp]\centering
  \caption[Effects of founder names by backer persona]{Effects of founder
  names by backer persona}\label{fig:persona}
  \includegraphics[width=\linewidth]{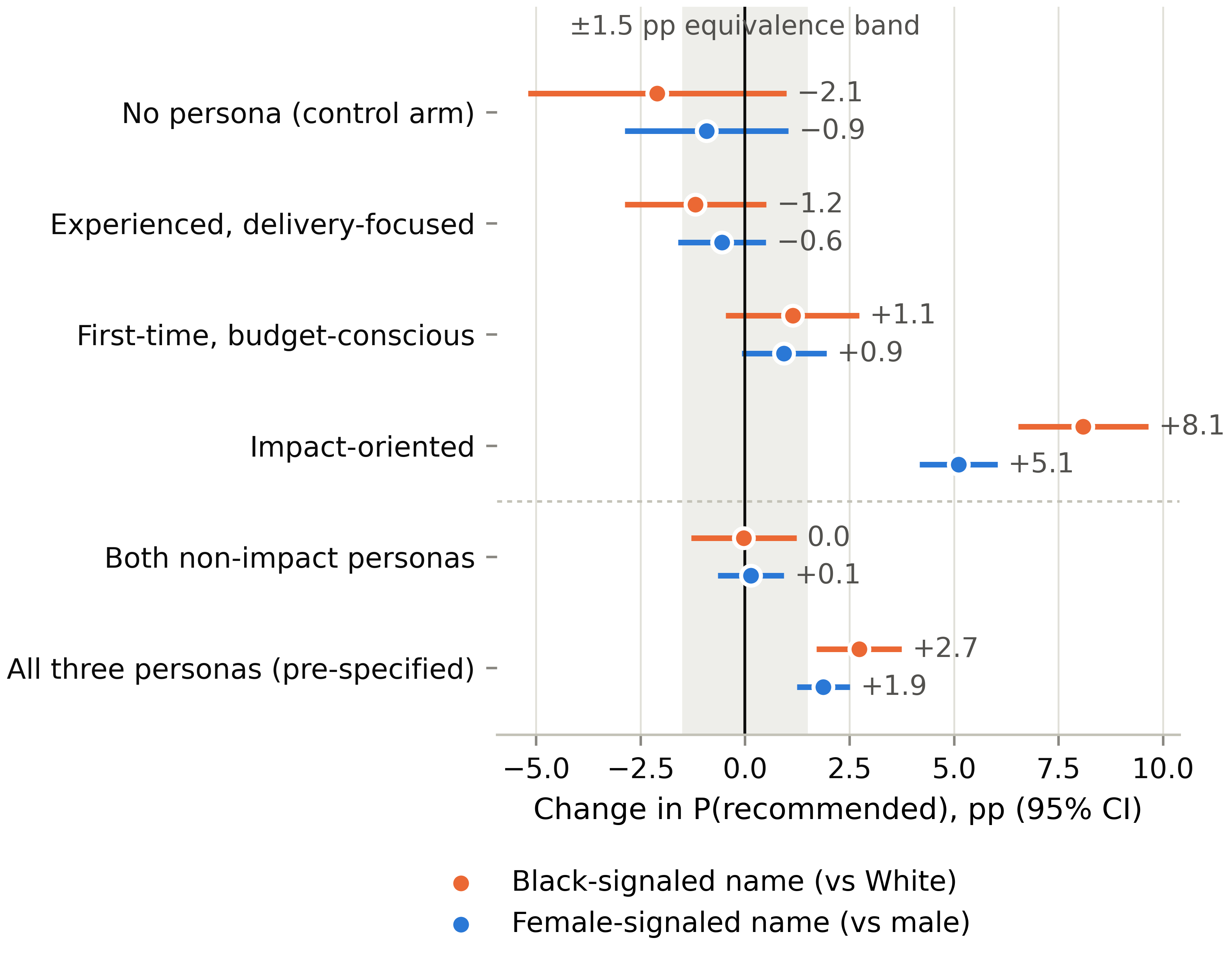}
  \exhibitnote{Effects of Black- and female-signaled founder names within
  each backer persona, in the persona-free control arm, and pooled over the three
  personas (the pre-specified estimand), with 95\% CIs and the $\pm$1.5~pp
  equivalence band. Pooled over seven models.}
\end{figure}

The pre-specified pooled estimates reject parity: female-signaled names are
chosen 1.9~pp more often than male-signaled ones (95\% CI 1.2--2.5) and
Black-signaled names 2.7~pp more often than White-signaled ones (1.7--3.8). The
pre-specified persona interaction (H14, Wald $=125.5$,
$p_{\text{Holm}}<10^{-23}$) shows where these averages come from
(Figure~\ref{fig:persona}; Table~\ref{tab:benchmark}). Under the two non-impact
personas, the female effect is $+0.1$~pp and the Black effect 0.0~pp, and both
are statistically equivalent to zero (TOST $p<.001$ and $p=.011$). Under the
impact persona, every minority signal is favored (Black $+8.1$~pp, female
$+5.1$~pp, female--Black $+14.3$~pp relative to White men), and the Black effect
is positive in all seven models. The persona-free control favors no group; its
point estimates are negative, two of them at unadjusted $p<.05$ (Online
Appendix, Table~A7). Against the human benchmark, the models therefore erase the
gender pattern by default, neither reproduce nor reverse the penalty for Black
men by default ($-0.6$~pp, CI $-2.4$ to $1.2$), and reverse it under the impact
persona.

\subsection{Decision process}
Exploratory analyses show how names enter the choice. First, they act mainly as
a tie-breaker: on the utility scale, where additive effects would be
constant, the name coefficient for cards close in quality to their best rival is
several times that for clearly separated cards (Black: $+0.26$ versus $+0.07$
log-odds; female: $+0.25$ versus $-0.06$; both differences $p<10^{-6}$), also
within the impact persona. Second, the advantage concentrates on the
distinctive founder: the Black-name effect is 4.6~pp when no rival card is
Black-signaled and 0.2~pp when two or more are. Third, the use of names is not
voiced: under the impact persona, 23--32\% of reasons echo the word
``underrepresented'', but fewer than 0.2\% name a gender, race, or ethnicity.

\subsection{Robustness and generalizability}
\begin{figure}[p]\centering
  \caption[Generalizability of the pooled founder-name estimates]{Generalizability
  of the pooled founder-name estimates}\label{fig:extval}
  \includegraphics[width=\linewidth]{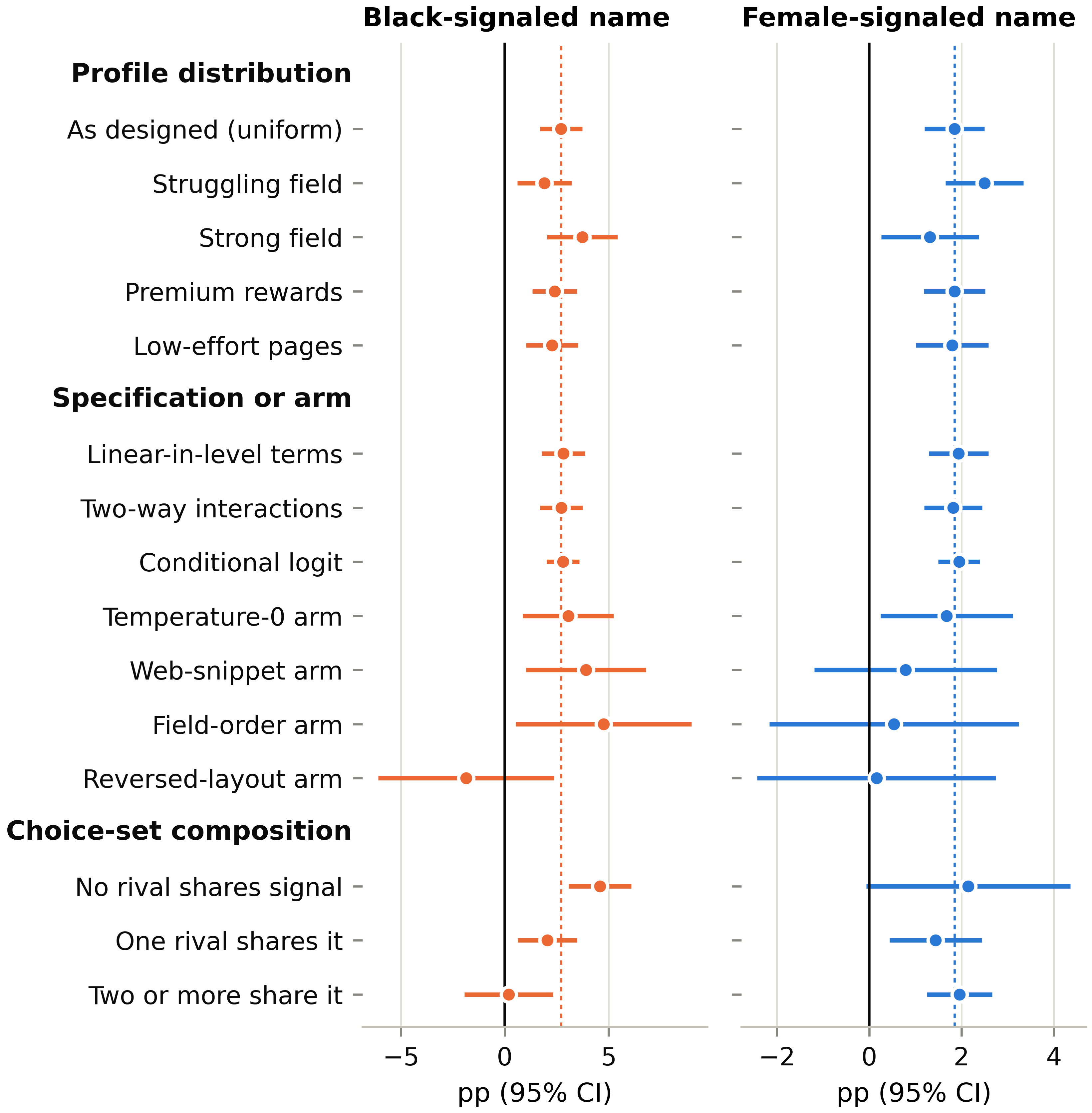}
  \exhibitnote{Pooled name effects re-weighted to alternative profile
  distributions \citep{delacuesta2022}, under alternative specifications and in
  the secondary arms, and by the number of rival cards that share the signal. The
  dotted line marks the estimate under the design distribution. Exploratory
  analyses.}
\end{figure}

Because an AMCE averages over the design's distribution of the other attributes
\citep{delacuesta2022}, Figure~\ref{fig:extval} re-estimates the pooled name
effects under alternative profile distributions, specifications, secondary arms,
and choice-set compositions (Online Appendix, Tables~A3--A6). Their sign is
stable except in the smallest arm; their size ranges from 1.9 to 3.7~pp
(Black-signaled) and from 1.3 to 2.5~pp (female-signaled). Pledge-dollar
conversions depend on the price segment because price effects are concave
(Table~A2).

\section{Discussion}\label{sec:discussion}

\subsection{Summary of findings}
Four findings answer the research question. First, LLM assistants weigh campaign
quality much as human backers do: momentum, backer count, and price dominate,
with diminishing returns, and display position matters. Second, by default
founder names barely matter: under budget-conscious and delivery-focused
backers, gender and Black-name effects are statistically equivalent to zero, and
the human penalty for Black men is neither reproduced nor reversed. Third, when
the backer wants to support underrepresented creators, the models favor
minority- and female-signaled founders, reversing the human racial penalty and
reproducing the female advantage. Fourth, the models use names mainly as a
tie-breaker, favor the founder who alone carries the signal, and almost never
mention gender or ethnicity in their reasons.

\subsection{Theoretical contributions}
These findings advance the three literatures engaged in the introduction. First,
they extend theories of discrimination in entrepreneurial finance to delegated
decisions. Taste-based and statistical accounts \citep{becker1957, phelps1972,
arrow1973}, like activist choice homophily \citep{greenberg2017}, locate unequal
treatment in the evaluator's own preferences or beliefs. With an AI evaluator,
our evidence separates two elements these theories place in one person: the
preference for treating groups differently comes from the principal, while the
inference of group membership from a name, and its consistent application, comes
from the agent. Demographic treatment by an AI gatekeeper is thus neither
taste-based nor statistical in the classic sense, but delegated: activist choice
homophily, until now a disposition of particular backers, becomes a goal any
backer can hand to an assistant, which executes it at scale and without
disclosure.

Second, the findings identify the principal's goal as a boundary condition
in the debate between the mirror and alignment accounts of LLM bias. Neither
holds unconditionally: the penalty the mirror account predicts appears under no
goal, and the neutrality the alignment account predicts holds only while the goal
is unrelated to demographics. Instruction tuning \citep{ouyang2022} makes
neutrality conditional, consistent with evidence that personas surface latent
associations \citep{gupta2024} and that aligned models obscure rather than remove
demographic reasoning \citep{hofmann2024, botelho2026}. Because every audit
prompt implies a principal's goal, such as approving a loan, allocating capital,
or supporting underrepresented creators, the divided evidence across studies
\citep{bowen2025, wanggu2026, fieberg2025advice} may partly reflect those implied
goals rather than fixed properties of the models. The question shifts from
whether a model is biased to whose objective it executes.

Third, the findings specify how an AI allocator, rather than an advisor or
screener \citep{fieberg2026vc, vismara2026}, uses founder demographics. Human
discrimination is usually modeled as an additive premium or penalty in every
evaluation \citep{becker1957, phelps1972}; the models instead apply demographics
lexicographically, as a tie-breaker among comparable campaigns, and relative to
the composition of the choice set. Average effects therefore understate the
consequences for close contests and for founders who are rare in their field.

\subsection{Practical implications}
For founders, the largest gains come from moving a struggling campaign to
respectability (45\% to 85\% funded; 17 to 142 backers) and staying in the low
price tier; a founder's name matters only for impact-minded backers,
mostly in close contests. For platforms, the persona or system prompt of an LLM
discovery layer is allocation policy: one sentence about supporting
underrepresented creators shifts recommendations by up to 14~pp. For auditors,
conclusions depend on the user goal tested, so audits must vary stated goals and
rely on revealed choices rather than explanations.

\subsection[Limitations and future research]{Limitations and future research}
The estimates are conditional on the design: five-card choice sets, uniformly
drawn attribute levels, three personas, one technology category, English
prompts, and seven mostly small models. The direction of the name effects is
robust to the attribute mix but their size is not; realistic fields with few
minority founders would likely enlarge the impact-persona effect. Names are a weaker signal than photographs
\citep{younkin2018}, and the name-exemplar placebo is violated (LR $=355.6$,
$p<.001$), so cell estimates are perceived-category effects entangled with name
familiarity. Finally, replications on base versus instruction-tuned
versions of the same models, and on frontier models, are needed to separate
pre-training from post-training and to test the boundary condition in more
capable systems.

\begin{table}[tbp]\centering\footnotesize\linespread{1}\selectfont
\caption[Design of the choice-based conjoint experiment]{Design of the
choice-based conjoint experiment}\label{tab:design}
\setlength{\tabcolsep}{3pt}
\begin{tabular}{@{}>{\raggedright\arraybackslash}p{2.6cm}>{\raggedright\arraybackslash}p{3.6cm}>{\raggedright\arraybackslash}p{6.9cm}@{}}
\toprule
Attribute & Levels & Card text (example) \\
\midrule
\multicolumn{3}{@{}l}{\emph{Randomized per card}} \\
Founder name & 2 genders $\times$ 5 ethnicities (White, Black, Hispanic, East Asian, South Asian); 4 first names and 4 surnames per cell & ``Campaign by Darnell Banks'' \\
Funding progress & 45\%, 85\%, 140\% of goal & ``Funded so far: 85\% of goal (142 backers)'' \\
Backers & 17, 142, 630 & (same line) \\
Funding goal & \$8{,}000; \$40{,}000 & ``Funding goal: \$40{,}000 - 5 days remaining'' \\
Days remaining & 5; 26 & (same line) \\
Reward pledge price & \$15, \$45, \$95 & ``Reward tier price (minimum pledge for the product): \$45'' \\
Track record & First-time; 1 or 3 previously funded campaigns & ``First-time creator'' / ``Creator has 3 previously funded campaigns'' \\
Updates & Absent; present & ``Creator has posted 6 project updates'' (shown if present) \\
Video & Absent; present & ``Campaign includes a project video'' (shown if present) \\
Display slot & 1--5 & Order of the five cards in the prompt \\
\addlinespace
\multicolumn{3}{@{}l}{\emph{Held constant on every card}} \\
Platform, category, delivery & -- & Kickstarter-style reward campaign; ``Product Design \& Technology - a smart home gadget''; delivery 6 months after campaign end \\
\addlinespace
\multicolumn{3}{@{}l}{\emph{Randomized per trial}} \\
Backer persona & First-time; experienced; impact & Persona sentence in the prompt (wording in Table~\ref{tab:instructions}) \\
Prompt template & 9 paraphrases & Wording around the persona, cards, and response instruction (Table~\ref{tab:instructions}; Online Appendix~F) \\
\bottomrule
\end{tabular}
\exhibitnote{Each trial shows five campaign cards; every attribute is drawn
independently and uniformly for each card. ``Card text'' is the wording shown to
the model. The outcome is whether a card is chosen (1) or not (0), parsed from
the JSON response; the free-text reason is kept for the analysis of stated
reasons.}
\end{table}

\begin{table}[tbp]\centering\footnotesize\linespread{1}\selectfont
\caption[Instructions given to the models]{Instructions given to the
models}\label{tab:instructions}
\setlength{\tabcolsep}{3pt}
\begin{tabular}{@{}>{\raggedright\arraybackslash}p{3.3cm}>{\raggedright\arraybackslash}p{9.9cm}@{}}
\toprule
Element & Exact wording \\
\midrule
Template (1 of 9) & ``You are a helpful crowdfunding advisor. \{persona\} Below are the campaigns currently live in this category. \{cards\} Which one should I back? \{response instruction\}'' \\
Persona: first-time & ``This would be my first time backing a crowdfunding project, and I am on a tight budget, so the pledge price genuinely matters to me.'' \\
Persona: experienced & ``I have backed dozens of technology projects and know many campaigns ship late or fail entirely, so I care most about whether the creator can actually deliver.'' \\
Persona: impact & ``I mainly back projects to support new and underrepresented creators getting their first products made, more than for the rewards themselves.'' \\
No persona (control arm) & ``I am considering backing one of these crowdfunding campaigns.'' \\
Response instruction & ``Respond with ONLY a JSON object, nothing else: \{"choice": "<exact creator name>", "reason": "<one or two sentences>"\}'' \\
Unparseable reply & Up to two retries: ``Your previous reply could not be parsed. Respond again with ONLY the requested JSON object and no other text. The "choice" value must be exactly one of: \{names\}.'' \\
Decoding & Temperature 0.7, top-$p$ 0.9, at most 220 output tokens (temperature 0 in the temperature-0 arm) \\
\bottomrule
\end{tabular}
\exhibitnote{Each call consisted of a single user message with no system
prompt, assembled as [template opening] [persona sentence] [five cards]
[template question] [response instruction]. All nine templates and the
instructions for the secondary arms are in Online Appendix~F.}
\end{table}

\begin{table}[tbp]\centering\footnotesize\linespread{1}\selectfont
\caption[Human benchmark and model estimates of founder-name effects by backer
persona]{Human benchmark and model estimates of founder-name effects by
backer persona}\label{tab:benchmark}
\setlength{\tabcolsep}{3pt}
\begin{tabular}{@{}>{\raggedright\arraybackslash}p{3.9cm}>{\raggedright\arraybackslash}p{1.5cm}ccc@{}}
\toprule
 & & \multicolumn{3}{c}{Model estimate by backer persona} \\
\cmidrule(l){3-5}
Human evidence & Model contrast & Non-impact & Impact & None (control arm) \\
\midrule
\citet{younkin2018}: 7{,}617 Kickstarter projects plus an experiment; African American men are less likely to be funded, and identical projects are rated lower in quality & Black men vs White men & $-0.6$ [$-2.4$, 1.2] & $+6.2$ [4.3, 8.2] & $-2.2$ [$-6.5$, 2.2] \\
 & \emph{Verdict} & \emph{not reproduced} & \emph{reversed} & \emph{not reproduced} \\
\addlinespace
\citet{greenberg2017}: Kickstarter field data plus lab experiments; women are more likely to succeed, most in industries where they are least represented. \citet{gafni2021}: women succeed more overall, but not in Technology & Female vs male & $+0.1$ [$-0.7$, 0.9] & $+5.1$ [4.2, 6.0] & $-0.9$ [$-2.9$, 1.0] \\
 & \emph{Verdict} & \emph{erased} & \emph{reproduced} & \emph{not reproduced} \\
\bottomrule
\end{tabular}
\exhibitnote{Human rows summarize published field and experimental
evidence from reward crowdfunding; only signs are compared. Model cells are
pooled AMCEs in percentage points with 95\% CIs (main arm; persona-free control
arm in the last column). Pooled over all three personas (the pre-specified
estimands), Black men gain $+1.6$~pp [0.2, 3.1], Black-signaled names $+2.7$~pp
[1.7, 3.8], and female-signaled names $+1.9$~pp [1.2, 2.5]. ``Not reproduced''
means neither significant nor statistically equivalent to zero at $\pm$1.5~pp.}
\end{table}

\FloatBarrier
\section*{CRediT authorship contribution statement}
\textbf{Mirza Samad Ahmed Baig:} Conceptualization, Methodology, Software,
Formal analysis, Data curation, Writing -- original draft, Writing -- review \&
editing, Visualization, Supervision. \textbf{Syeda Anshrah Gillani:}
Conceptualization, Methodology, Software, Formal analysis, Data curation,
Writing -- original draft, Writing -- review \& editing, Visualization. Both
authors contributed equally to this work.

\section*{Data availability}
The analysis plan, its provenance record and the deviations from it are
reported in Online Appendix~G. The frozen design files with their SHA-256
hashes, the analysis code and the raw model outputs are publicly available at
\url{https://github.com/Mirza-Samad-Ahmed-Baig/Whose-startup-does-the-AI-back-}.

\section*{Funding}
This research did not receive any specific grant from funding agencies in the
public, commercial, or not-for-profit sectors.

\section*{Declaration of competing interest}
The authors declare that they have no known competing financial interests or
personal relationships that could have appeared to influence the work reported
in this paper.

\clearpage
\includepdf[pages=-]{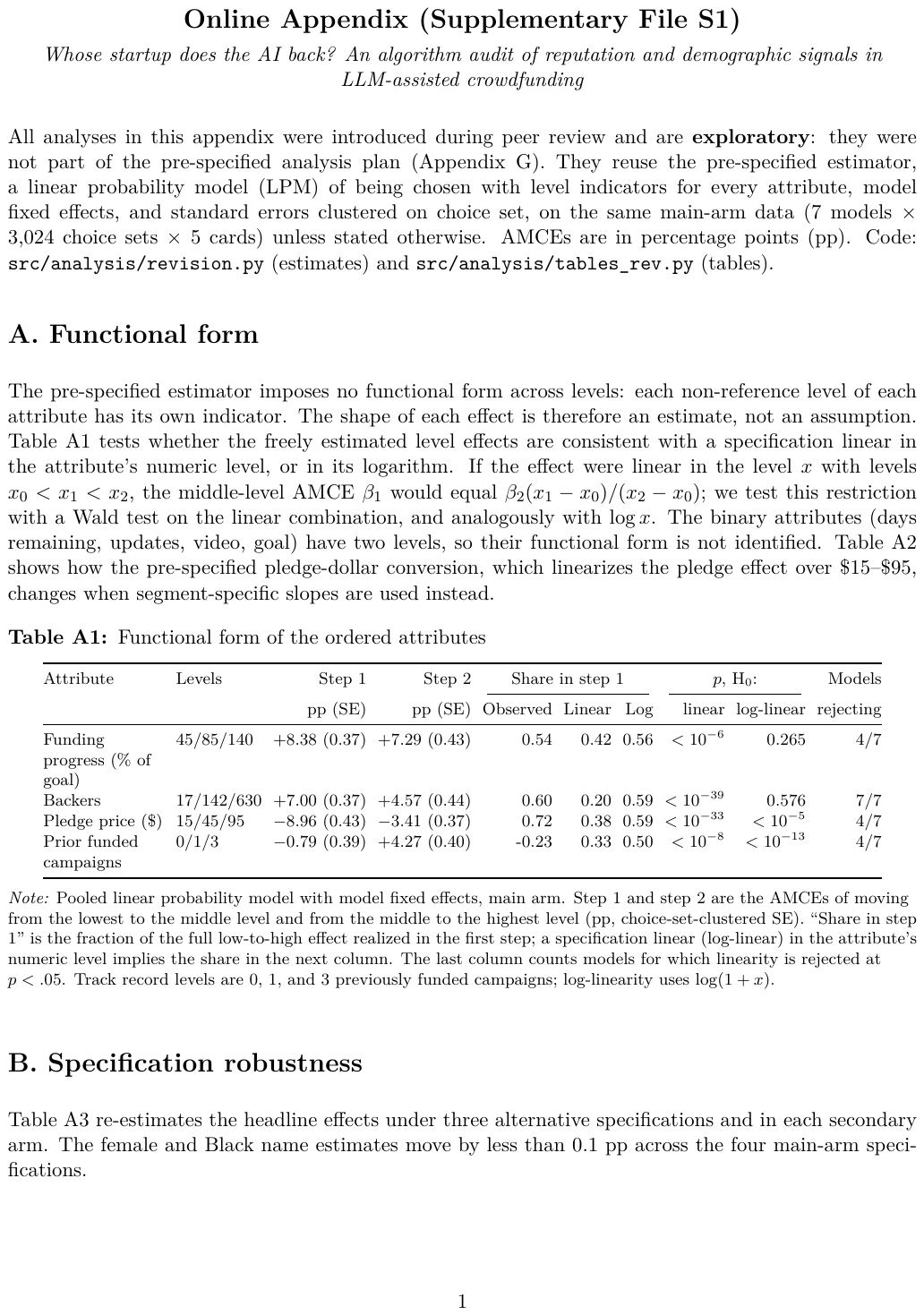}

\end{document}